# Service Pet Robot Design: Queer, Feminine and Sexuality Aspects

Anna-Maria Velentza[1,2,3] and Antigoni Tsagkaropoulou[3,4]

[1]School of Educational & Social Policies, University of Macedonia, GR, [2] LIRES Robotics Lab, University of Macedonia, GR

[3] Fulbright Scholar Greece, [4]UCLA School of Art & Architecture, Department of Design Media Art, USA

*Abstract*— The integration of robots and AI in society raises concerns about discrimination and biases mostly affecting underrepresented groups, including queer and feminine figures. Socially assistive robots (SAR) are being used in a variety of service and companion roles, following social norms during their interaction with humans and seem to be beneficial in many roles, such as the pet therapy robots. To promote inclusion and representation, robot design should incorporate queer and feminine characteristics. As a response to these concerns, a pet robot called BB was designed using a multidisciplinary and inclusive approach. BB was presented in a queer architecture and aesthetics environment, emphasizing aspects of techno-touch, vulnerability, and sexuality in human-robot interactions. The audience's perception of both the robot and the female researcher was evaluated through questionnaires and focus groups. This study aims to explore how technology and design can better accommodate diverse perspectives and needs in the field of SAR.

*Index Terms*—human robot interaction, robot design, queer, feminine, evaluation

## I. INTRODUCTION

The integration of robots into society gives rise to social, ethical, and even legal dilemmas. It is well-known that discrimination and bias are inherent challenges in numerous AI applications. This leads us to question how these issues impact the underrepresented groups, queer, and feminine figures and whether they are considered during the development and utilization of robotics and AI[1]. Surprisingly, the exploration of how machines affect them remains largely unexplored. Only a limited number of works in the literature delve into LGBTQ+ concerns in the design of robots and artificial intelligence (AI) [1]. The categorization of AI applications into clear and distinct groups, often based on binary distinctions, is dependent on ways of creating knowledge that are ingrained in systemic authority and mythological frameworks[2]. These frameworks promote a narrow perspective of human identity, adhering to heteronormative and liberal ideals, which consequently restricts and confines our understanding of humanity [2]. Apart from the under representation of people interacting with robots, there are still major unsolved problems with gender biases and underrepresented groups in science[3], [4]. Evidences show that increasing femininity in a person's appearance, decreases the likelihood of being perceived as being a scientist [5].

Socially assistive robots (SAR) refer to robots that aid or support to humans in social interactions. These robots prioritize meaningful and productive interactions by facilitating effective and intimate engagement with human users, enabling assistance and tangible advancements in areas such as recovery, rehabilitation, education, and other domains[6].

The success of human-robot interactions is heavily dependent on the robots' proficiency in communication across various modalities, encompassing both verbal and nonverbal means [7]. Emotion plays a central role in these forms of communication, expressed through facial expressions, speech, colors, and full-body motion [8]. Particularly, the robots' expressive body language significantly influences various aspects of interaction, such as shaping humans' attitudes towards them, evoking emotional arousal, and impacting the perception of the interaction itself [9], [10], [11]. Moreover, the robots' speech, both auditory and semantic, holds a crucial position in expressing emotional behavior during human-robot interactions [12]. By mastering these elements of communication, robots can foster meaningful and effective engagements with humans, leading to more natural and beneficial interactions in various domains.

Studies have shown that individuals tend to ascribe moral norms to robots in a manner analogous to their evaluations of human beings[13]. This phenomenon is indicative of a process wherein the same social and emotional characteristics that influence human-human interactions also impact robot-human interactions[14]. Additionally, over an extended period, pet therapy has been acknowledged for its considerable emotional benefits. Recent investigations have demonstrated that robotic pets yield analogous positive effects, while circumventing the potential drawbacks associated with conventional live pets. Consequently, robotic pet therapy presents a viable and promising alternative to traditional pet therapy approaches[15].

## II. RELATED WORK

The design of human-like robots accompanying and serving us in everyday activities demands a deep understanding of the human body, humanoid and animistic cues, and to go beyond heterosexual gender norms. Treusch suggested performative demonstrations for people to experience and co-design the humanoid robot companions [16]. Joo [17] suggests the notion of an Asian robot, embracing the idea that it opens the door to exploring alternative perceptions of race that go beyond mere association with human form. Following his lead, we support the idea that a queer robot can embrace queer perspective in society. Moreover, Shildrick [18] examines how the unique practices and settings of dementia care can be reshaped and challenged by the introduction of technological and intelligent assisted aids, and particularly zoomorphic robots, which are progressively augmenting traditional human assistance.

### III. Present Study

In the present study, we focus on the first part of the lecture performance 'BB the pet robot', where we design and implement from a queer and feminine perspective a service pet robot prototype costume, and we evaluated both the female lecturer who was presenting it and the robot design with Likert scale questionnaires and focus group. The artist and second author wore the robot costume and acted as it.

### IV. BB the Pet Robot

*a) Robot Design*

The robot called BB, is human sized, as shown in Fig.1, with humanoid and animal characteristics. The morphology follows all three major design concepts of SAR, the cute companionship, the human mimic and a futuristic machine [19]. The robot's fur is fluffy and very soft to touch. The increasing prevalence of companion and empathy robots in assistive applications introduces the concept of "techno-touch," which encompasses both the physical aspect of touch and its metaphorical representation for emotional connection. In both senses, the act of touching and being touched should be considered in relation to vulnerability, not as a precarious state but as a gateway to an imagined realm of inseparable interconnections, where the distinction between self and other loses significance [18]. Moreover, BB has a light blue color, since blue is linked with stress reduction, positive reactions such as excitement and happiness, while it is considered attractive, inflating people's attention without affecting their memory and acceptable throughout cultures [20], [21].

BB has four paws and a tablet as a face with expressive facial expressions changing according to the context of the speech. The variation of facial expressions were the size and movements of the eyes and mouth. The voice is both machinery and humanoid, non-binary and in accordance with the appearance. Moreover, the robot throughout the presentation was making robotic sounds and it was doing slight movements even when it was not competing a task, to be perceived as alive. The robot's facial expressions, voice and storytelling are cute, following the evidence of the importance of cuteness in human-robot interaction and especially for service robots to be considered more likable[22], [23].

Additionally, we focused on creating a queer aesthetics indoors lecture environment for the presentation of BB, which differs from the heteronormative space. The strength of an exhibit or installation lies in its transience and apparent lack of practical purpose. The very "uselessness" of these installations leads to unforeseen revelations, compelling individuals to pause, ponder, and reconsider their conventional interactions with architecture[24]. Thus, we combined the usefulness, the existence of a table for the laptop and the monitors and a curtain to hide the robot before the reveal with the "uselessness", covering the table with shiny fabric and using a sequin and spangle curtain, as shown in Fig.1, C. The act of exposing and sharing what is typically private or concealed, and more appropriately, making it visible (sometimes quite literally), serves as a disruption and subversion of the conventional domestic space. This process can be seen as a form of "queering," challenging traditional norms and expectations associated with private areas [24]

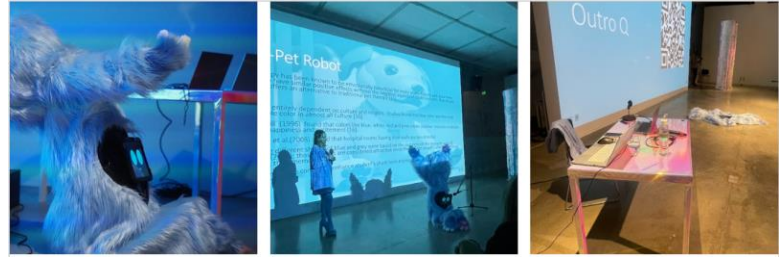

FIG.1: FROM LEFT TO RIGHT, A) THE ROBOT WITH THE TABLET AS FACE, B) THE LECTURER WITH THE ROBOT IN THE STAGE, C) THE LECTURE ENVIRONMENT

*b) Participants*

The lecture performance was promoted using printed flyers and electronic communication channels, specifically through university e-mail lists. The promotional materials included the names of both authors, along with their respective professional expertise, to establish credibility and context. In adherence to research methodology principles and to prevent potential expectation biases [25], explicit details regarding the robot's functionalities or appearance were deliberately omitted from the promotional content. Instead, the audience was informed solely about the presence of a pet robot, without any specific information on its capabilities or visual attributes. 60 people from the audience agreed to fill in the questionnaire and 20 out of them participated in the focus group the next day. Their age range was 21-72, 30 identified as she/her (50%), 12 as he/him (20%), 15 as they/them (25%), and the rest preferred not to mention.

*c) Procedure*

The lecture performance spanned across two days, each held at distinct venues. At the beginning, the lecturer, and first author delivered a comprehensive presentation with the aid of PowerPoint on SAR, elucidating their interactions with individuals in various tasks and expounding on their communication channels, akin to the Introduction section, and also by highlighting papers from her personal research. Following the lecture's conclusion, the lecturer revealed BB, the robot, by drawing back the curtain depicted in Fig. 1,C. Subsequently, the robot introduced itself to the audience, employing endearing sounds, robotic gestures, and facial expressions to accompany its storytelling.

During this phase, the lecturer expounded upon the practicality of employing pet robots in service-oriented activities within the realm of human-robot interaction. Additionally, the rationale behind the robot's specific appearance, encompassing its fluffy texture and color, was elucidated (Fig.1, B). To engage the audience further, the robot approached them, allowing individuals to interact by petting it.

Following this interaction, the lecturer requested the robot to return to the stage and subsequently commanded it to perform a dance by uttering the prompt, 'BB, dance.' This demonstration showcased the robot's capability to respond to verbal commands and execute dynamic movements, underscoring its potential as an interactive and entertaining technological entity. Following the feminism critiques the pornographic portrayals of femininity, and the prevailing sex robotics trends as intensifying outdated gender stereotypes to an extreme level, BB performed a sensual dance accompanied by a song. According to Kubes [26], the notion of robot sex as a sexist representation influenced by toxic masculinity. Advocating from a new materialist, sex-positive, and queer standpoint, freeing robots from the compulsion to imitate the human body with great precision could serve as a remedy to the discomfort and aversion often experienced when encountering humanoid but not entirely human-like replicas [26]. After 2 minutes of dance, the lecturer commanded the robot to stop and presented a QR code at the main screen, asking the audience to fill in the questionnaires to help them with their research.

*d) Measurements*

To evaluate both the lecturer and the robot we used 5 Likert Scale Aesthetic valence questionnaire [27], using the same questions to evaluate them both [28]. The questions were regarding their appearance, i.e., 'lovely, attractive', the uncanny valley, i.e., 'scary, creepy', the perceived personality i.e., 'cheerful, boring', their functionality i.e., 'functional, trustworthy'. Furthermore, we added two provocative, based on gender stereotypes items, 'slutty', and 'affectionate'. Additionally, to the questionnaires, the next day of the performance a focus group/ critique section took place.

*e) Results*

As positive evaluation we sum up the replies scoring 4 and 5 in the Likert scale (agree and strongly agree) and then we calculate the percentage out of the total replies per question (N=60). The audience highly evaluated the robot and the lecturer in terms of trustworthiness (46.66%, N=28 for the robot and 70%, N=42 for the lecturer) and effectiveness (50%, N=30 for the robot, and 80%, N=48 for the lecturer). Additionally, they both perceived as attractive in terms of physical appearance (76.66%, N=46 for the robot, and 86.66%, N=52 for the lecturer), while they were both not perceived as uncanny, receiving low scores in the corresponding items, creepy and scary (25% for the robot and 13.33% for the lecturer).

It is worth mentioned that enough people evaluated the lecturer as 'slutty', with 28.6% scoring 4 and 5 in the Likert scale and 39.3% not being suer, scoring 3, neither agree nor disagree. Similarly for the robot, 50% evaluated as 'slutty', while 35.7% scored 3 in the Likert scale. Moreover, the robot was noticed to have a higher score evaluated as 'affectionate', 85.7%, in comparison with the lecturer, 46.4%.

During the focus group the participants agreed that they couldn't believe that the lecturer was a real researcher and scientists and that she looked like an actor playing the role of the researcher. Moreover, they criticized the existence of the sensual dance, with all men who participated to pointed out that it was unnecessary, while female and queer participants mentioned that they understand that there is a point in the existence of that dance, but maybe they should try to explain it verbally to the audience.

## V. DISCUSSION & CONCLUSIONS

The primary objective of this study is to establish guidelines for the design of SAR that embrace inclusivity and queerness. This advancement builds upon the incorporation of vulnerability and feminine characteristics in SAR design, which contributes to providing support and eliciting positive emotional responses from users. Furthermore, the inclusion of these aspects prompts critical discussions, both at the societal and individual levels, pertaining to the dynamics of human-robot interaction, as well as the evolving relationships with one's own body, behavior, and sexuality in the context of an increasingly technologically driven world.

By infusing SAR with qualities that evoke vulnerability and femininity, the aim is to create robots that offer empathetic and nurturing attributes, promoting a sense of emotional connection and well-being among users. This approach challenges traditional notions of robotic design and explores alternative paradigms of human-robot interaction that are sensitive to individual diversity and needs.

Consequently, the introduction of inclusive and queer SAR engenders contemplation and dialogue on broader societal implications, such as public perceptions of robots, privacy concerns in human-robot interactions, and the influence of smart technologies on shaping human behavior and attitudes towards their own sexuality and identity. Ultimately, it is envisioned that this research will contribute to the development of more empathetic and socially aware SAR that can positively impact various user groups and foster a more inclusive and accepting technological society.

BB performed social characteristics and the outcome of the questionnaire showed that the ethical and moral considerations governing human behavior are extended to encompass the behavior of the robots as well. The social and emotional factors that influence the dynamics of human interactions play a significant role in shaping how individuals perceive and interact with robots[13], [14] and the audience evaluated BB and the lecturer with similar moral norms.

Our findings are in line with [5], where the female characteristics were not perceived as scientific. In our case, the identity and research focus of the lecturer were prominently highlighted across various mediums, including the advertisement flyer, the introductory segment of her lecture, and even within the literature used for the presentation. Despite this comprehensive visibility of her credentials, the audience expressed scepticism regarding her research and authenticity, conjecturing that she might be an actress portraying the role of a scientist. This perception appears to be associated with the lecturer's perceived attractiveness and charm, which garnered significant attention and evaluation from the audience.